\documentclass[fleqn,10pt]{wlscirep}
\usepackage[utf8]{inputenc}
\usepackage[T1]{fontenc}
\usepackage{amssymb}
\usepackage{amsmath}
\usepackage{makecell}
\usepackage{booktabs}
\usepackage{array,multirow,graphicx}
\usepackage{float}
\usepackage{graphicx}
\usepackage{subcaption}

\title{Angular momentum analysis on Karate roundhouse kicks: a longitudinal case study}

\author[1,*]{Jan C. L. Lau}
\author[2,]{Christian Mele}
\author[2,]{Jonathan Feng-Shun Lin}
\author[1,2]{Katja Mombaur}
\affil[1]{BioRobotics Lab, Optimization and Biomechanics for Human-Centred Robotics (HCR), Institute for Anthropomatics and Robotics (IAR), Karlsruhe Institute of Technology, 76131 Karlsruhe, Germany}
\affil[2]{Canada Excellence Research Chair in Human-Centred Robotics and Machine Intelligence, Systems Design \& Mechanical and Mechatronics Engineering, University of Waterloo, Waterloo, N2L3G1, Ontario, Canada}

\affil[*]{jan.lau@kit.edu}


\begin{abstract}
Human gait in reality extends beyond straight-line walking, with some situations even requiring drastic back-and-forth rotations. A smooth and stable execution may seem intuitive, but the underlying mechanics remains unknown. The Karate roundhouse kick may be an extreme case of exhibiting dynamic back-and-forth rotations, but investigating the angular momentum (AM) management can potentially inform stability analysis in dynamic human motions and smoother gait in robots and exoskeletons. This paper introduces two new AM-based measures and analyzes AM-related variables to study the target-less retractable back-leg Karate roundhouse kick. The purpose is to understand the underlying AM management, analyze the differences between stable and unstable kicks, and investigate how these variables and measures change over time with improvement. A one-year longitudinal study was conducted with the first author as a Karate student, and a Karate instructor was also recruited for one session as an expert, whose data is used for comparison. Results show that unstable kicks have higher peak total AM before Strike and smaller braking peak after Strike. The proposed AM-based measures are able to identify the cause of some unstable kicks, though no clear distinction could be made between stable and unstable kicks since kicks can be unstable for different reasons. Nonetheless, the proposed measures can be applied as performance measures to analyze other types of dynamic motions. To incorporate them as stability criteria, however, it is recommended to also consider their coordination with time, kinematics, and center of pressure.

\end{abstract}
\begin{document}

\flushbottom
\maketitle

\thispagestyle{empty}

\section*{Introduction}
Rotating back and forth dynamically while performing activities of daily living (ADL) occurs naturally and is often executed stably in healthy adults. In older adults with decreased mobility, the same rotations during ADL can increase the risk of falling. In robotics, it is of interest to make robots and exoskeletons walk more human-like for better social acceptance and human-robot interaction. Therefore, understanding the proper rotation and angular momentum (AM) management can not only advance knowledge on dynamic human movement, but also provide insights to improve robot and exoskeleton control.

Effective rotation and AM management are often found in sports: conservation of AM is utilized to maximize spin velocities in figure skating \cite{figure_skating}, and tuck diving athletes must strategically transfer AM to produce twisting rotations given that AM is conserved during flight phase \cite{tuck_diving}. In martial arts, executing kicks with spins require balance, stability, and practice. Even though rotations in sports are considered more extreme than those in human gait, it is believed that the analysis could inform rotation stability during walking. After all, improved stability and whole-body control is a byproduct of martial-arts training, since Tai Chi practitioners consistently exhibit lower peak horizontal and vertical velocities at the shoulder, hip, and knee when being perturbed at the hips, shoulders, and arms \cite{sever_et_al_taichi}, and Judokas exhibit higher chances of motionless limb responses and absence of near-falls when perturbed \cite{betelli_et_al_judo_perturbation}.

The roundhouse kick is a challenging kick, where the person first lifts the knee ("chamber") while turning the supporting foot and body in a semi-circular motion, then extends the knee to strike a target with the lower part of shin (for harder strikes) or top of the foot (for gentler taps). To further improve execution stability, there exists a "target-less retractable" variation, where the person is required to stop on their own without external forces acting on them, then retract the kick leg to the starting pose at the end (see kick phases\cite{kim_et_al_tkd_rhk, gavagan_sayers_rhk} in Fig. \ref{fig:roundhousekick}). Since this kick requires extensive AM management for generating power and speed, it is very dynamic especially about the vertical axis. The roundhouse kick is commonly practiced in Karate and Taekwondo. They look similar with continued training and experience, but each martial art has a different focus in early education. In general, Taekwondo emphasizes speed and being light on the feet to allow for kick combinations, while Karate focuses more on power and maintaining a stable stance.

Various studies on the roundhouse kick were conducted, including and not limited to understanding Taekwondo roundhouse kick performance \cite{li_et_al_tkd_rhk, huang_et_al_tkd_rhk, thibordee_prasartwuth_tkd_rhk, estevan_et_al_tkd_rhk}, deriving critical Taekwondo roundhouse kick performance factors \cite{bercades_et_al_rhk_performance}, investigating kinematic differences in performing the kick with the preferred and non-preferred leg in elite Taekwondo athletes \cite{tang_et_al_tkd_rhk}, comparing roundhouse kick kinematics across different disciplines of martial arts \cite{gavagan_sayers_rhk, diniz_et_al_rhk} or against other kicks \cite{kim_et_al_tkd_rhk, vagner_et_al_fk_vs_rhk}, and analyzing the effects of Steady Motion Fitness \cite{abonyi_et_al} and neuromuscular electrical stimulation \cite{makronssios_rhk} on roundhouse kick performance. In contrast, only a few studies on the balance and stability of Karate motions were found. One study investigated the dynamic balance of Karatekas performing a predefined sequence of motion, with variables of interest including center-of-mass (COM) kinematics and angular motion of upper and lower limbs \cite{zago_et_al_dynamicbalance_karate_rhk}. Another study investigated roundhouse kick stability by analyzing center-of-pressure (COP) path length and distance from origin \cite{vando_et_al_karate_rhk}. Meanwhile, one study calculated balance and stability of a double side kick based on the horizontal and vertical position of the subject's COM and support leg toe and heel markers \cite{hoelbling_et_al_doublesidekick_balance}. Among the aforementioned literature, the target-less variation was only collected as one of two kick conditions in \cite{quinzi_et_al_rhk_neuromuscularcontrol}, but neither AM nor the stability aspect of the back-and-forth rotation was investigated. Executing this variation demands more stability given the change in rotation speed, AM, and direction, yet the back-and-forth rotations resemble those found in human walking.

AM is found to be an indicator of balance control when walking \cite{popovic_et_al_iros_2004, popovic_et_al_icra_2004, herr_popovic_2008, begue_et_al_2019, begue_et_al_2021}. Begue et al. analyzed the segmental AM with respect to (w.r.t.) the whole-body AM as a percentage \cite{begue_et_al_2021}, whereas Popovic et al. used principal component analysis (PCA) \cite{popovic_et_al_iros_2004, popovic_et_al_icra_2004, herr_popovic_2008}. In sports, AM is found to be an important characteristic in slackline balancing. In a study comparing experts and beginners, the results show that lower values in normalized AM and COM acceleration indicate less recovery movements and better stability in individuals \cite{kevin_slackline}. In robotics, Aller et al. identified AM as a more promising variable than the Foot Placement Estimator \cite{fpe}, Capture Point \cite{cp}, and Zero Moment Point \cite{zmp, zmp_sardain_bessonnet} when explaining the instability that happened to the bipedal humanoid robot REEM-C during walking \cite{reemc_eurobench_walking_stability}.

As mentioned, the target-less retractable back-leg roundhouse kick is considered an extreme scenario of back-and-forth rotations found in human walking. Even though the eventual goal is to develop a stability measure based on AM, one must first analyze AM behaviour in such scenarios. Therefore, this paper uses AM and proposes two new AM-based measures to understand the role of AM management in the back-and-forth rotations and analyze AM differences between stable and unstable kicks. Mele et al. stated that longer practice is needed to achieve expert-level neuromuscular coordination in Karate motions \cite{mele_et_al_2026}, so it is also of interest to study how AM management evolves with more training. We hypothesize that 1) the values of AM-related variables in stable kicks would increase with more Karate training, 2) unstable kicks have higher total AM values than stable kicks, 3) the kick leg AM orthogonal component (AMO) in stable kicks are below a certain threshold, and 4) the kick leg AM aligned component (AMA) in stable kicks is always positive leading up to the Strike. To the authors' knowledge, this is the first paper to analyze AM of roundhouse kicks and introduce AMA and AMO.

\section*{Methods}

\label{methods}
The first author (female, age 25, \(1.68m\), \(55.44kg \pm 1.86\)) participated as a Karate student and trained 3-4.5 hours per week. The second author (male, age 25, \(1.83m\), \(87.6kg\)), who is also a black belt Karate instructor with 12 years of training, participated as an expert. Both practice Goju-Ryu Karate.

\subsection*{Experiment Protocol}
\label{methods_experiment}
\begin{figure}[hbt!]
    \centering
    \includegraphics[width=\linewidth]{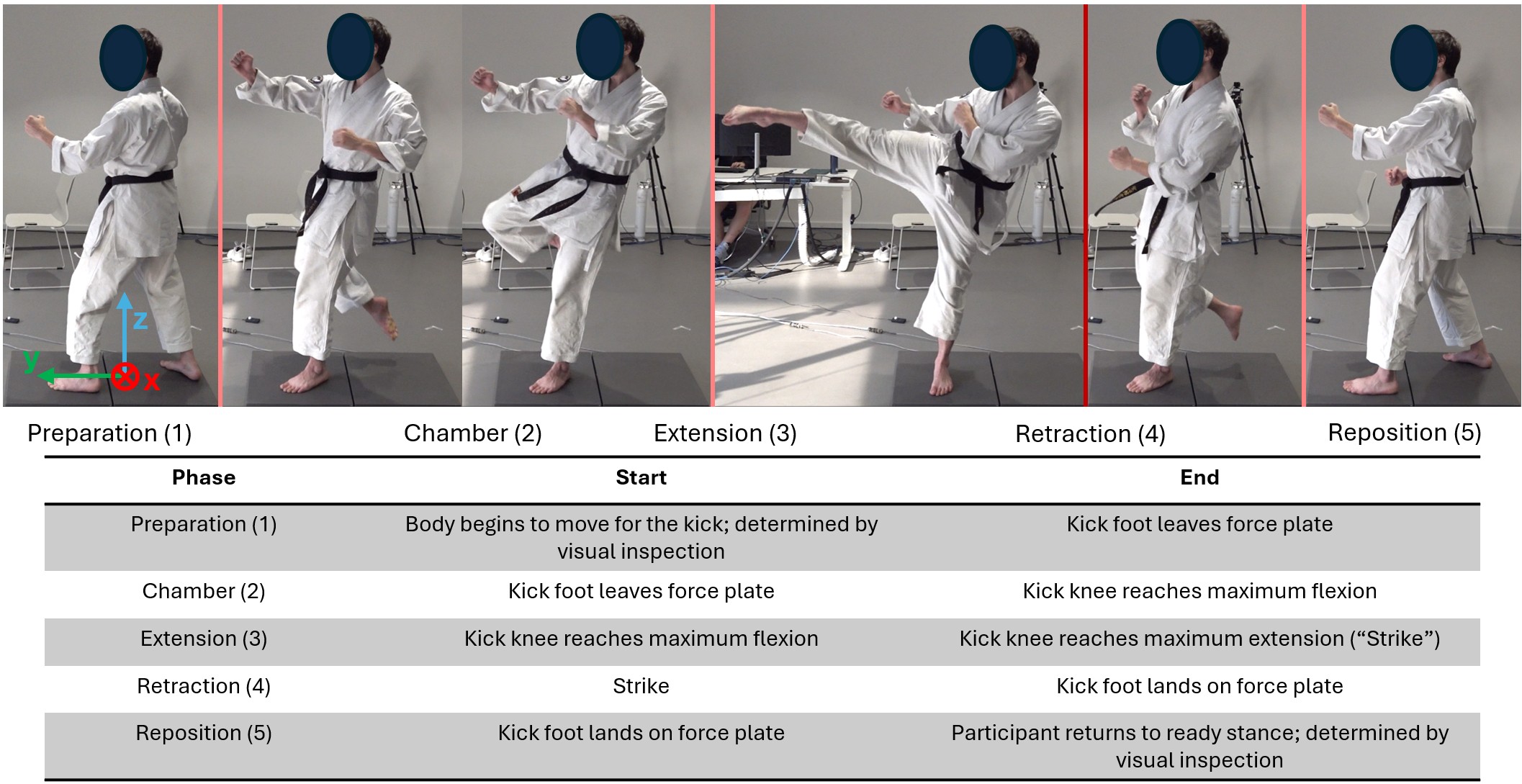}
    \caption{Sequential illustration of the Karate right back-leg roundhouse kick, with indication of start and end instances of all phases. This figure demonstrates the setup with force plates but without motion capture markers. The darker red line represents the Strike instance. Informed consent was obtained to publish the images above in an online open access publication.}
    \label{fig:roundhousekick}
\end{figure}

Eight sessions were collected with the student for one year, with each collection showing belt development over this time. A duration of one year was chosen because four months was deemed insufficient for Karate novices to develop expert-level characteristics \cite{mele_et_al_2026}. A single session was also conducted for the expert to obtain data for comparison. In each session, the participants performed ten target-less right back-leg roundhouse kicks. The five phases of the roundhouse kick are summarized in Fig. \ref{fig:roundhousekick}. “Right back-leg” means that the right leg is the kick leg and the ready stance involves the kick (right) leg being behind the left (support) leg (see Fig. \ref{fig:roundhousekick} for details). Target-less means the participants are responsible for stopping the kick and rotation upon knee extension by themselves, in contrast to kicking a target that stops the rotation.

The participants are not required to kick at a fixed height nor with full power, but are instructed to prioritize stability over kick height and speed, and must retract the kick to initial position. It is not required to kick with full power for safety reasons given the absence of a physical target, and the last requirement is chosen for its large amount of rotation about the vertical axis and higher demand on stability. A 12-camera Vicon Vantage system and two Bertec force plates were used for data collection. Fig. \ref{fig:roundhousekick} shows the interaction with the force plates, but the actual experimental setup includes passive markers placed on the participants, either on skin or tight clothing (never a Karate uniform (\textit{Gi})). A modified Istituto Ortopedico Rizzoli (IOR) full-body marker set with 53 markers is used \cite{modified_ior}. These data are part of a larger Karate dataset and are collected at the HCRMI Motion Capture Lab with ethics clearance from the Ethics Board of the University of Waterloo (REF.44013). The experiments were performed in accordance with their guidelines, and informed consent was obtained from both participants on experiment participation and publication of information/images in an online open access publication.

\subsection*{Post-processing}
\label{methods_postprocessing}
The Vicon Nexus software was used for data cleaning and exporting marker and force plate data. The Vicon Procalc software was used for redefining kinematic variables from marker placement. Joint kinematics and angular momenta (trunk, arms, legs, and total) were generated with an in-house pipeline, which takes Vicon c3d and vsk files as input and outputs kinematic variables. This pipeline considers de Leva anthropometric values \cite{deleva}, is written in Python \cite{python}, and uses ezc3d \cite{ezc3d} and biorbd \cite{biorbd}. For each kick, the start of Preparation and end of Termination are visually identified through video data from a Karate perspective, whereas the remaining phases were identified using joint angle and force plate data.

Per visual inspection of the motion capture video data, all kicks from all sessions were first labeled as “stable,” “marginally stable," and “unstable” based on the input from the Karate expert. A kick is considered stable if the person exhibits whole-body control and completes the full motion as planned, whereas a kick is considered unstable if the person had to jump or take a step for balance recovery. The purpose of doing so is to provide a baseline reference. The full criteria on how a stable kick is defined from a Karate standpoint can be found in the Supplementary Note S1 online.

\subsection*{Proposed Angular Momentum Measures and Variables Analyzed}
\label{methods_measures}
As mentioned, this paper proposes AMA and AMO as measures for AM analysis. All AM variables utilized and analyzed in this paper are w.r.t. fullbody COM.

Sharing the same unit as AM and with \(\vec{H} = [H_x, H_y, H_z]^T\), AMA describes the magnitude of a body part’s AM in the direction of total AM. From a mathematics perspective, it is the scalar projection of the body part's AM on the total AM, normalized to the L2-norm of total AM. This variable is believed to be more descriptive than expressing body part AM as a percentage of total AM. A positive AMA value means the body part’s AM is in the same direction as total AM, and vice versa (see Eq. \ref{eqn_ama}). \(\vec{AMA}\) is the AMA vector of the body part, which is the \(AMA_{bodypart}\) multiplied by the unit vector of the total AM \(\hat{H}_{bodypart}\). Note that this is not the first instance of computing dot products of two vectors for kick analysis, since Kim et al. defined the inter-joint coordination (IIC) index as the dot product between 3D unit vectors of the hip and knee joint angular velocities \cite{kim_et_al_tkd_rhk}. However, IIC is only limited to angular velocities and focuses on motion plane coincidence.

\begin{equation}
    \begin{split}
        AMA_{bodypart}(t) = \frac{\vec{H}_{bodypart}(t) \cdot \vec{H}_{total}(t)}{||\vec{H}_{total}(t)||}\\
        \vec{AMA}_{bodypart}(t) = AMA_{bodypart}(t) * \hat{H}_{total}(t)
        \label{eqn_ama}
    \end{split}
\end{equation}

AMO is the magnitude of a body part’s AM orthogonal to total AM to describe the body part’s influence on changing the direction of total AM (see Eq. \ref{eqn_amo}). This variable is analyzed since it is suspected that AMA alone does not provide a sufficient picture on how AM influences stability in a very dynamic motion like the roundhouse kick. AMO also shares the same unit as AM.

\begin{equation}
    \begin{split}
        \theta(t) = arccos\left( \frac{\vec{AMA}_{bodypart}(t)\cdot \vec{H}_{bodypart}(t)}{\lVert \vec{AMA}_{bodypart}(t) \rVert \lVert \vec{H}_{bodypart}(t) \rVert} \right) \\
        \vec{AMO}_{bodypart}(t) = \vec{H}_{bodypart}(t) - \vec{AMA}_{bodypart}(t) \\
        AMO_{bodypart}(t) = \lVert \vec{H}_{bodypart}(t) \rVert * sin(\theta(t))
        \label{eqn_amo}     
    \end{split}
\end{equation}

Previous works have analyzed segmental AM contribution using PCA \cite{popovic_et_al_iros_2004, popovic_et_al_icra_2004, herr_popovic_2008} and percentages \cite{begue_et_al_2021}. The rationale for the described approach is to obtain additional information that percentages would not reveal while keeping the calculations simple for 3D analysis (applicable to AMA and AMO). While PCA is a commonly-used method for visualizing multivariate data and identifying influence, the emerging principal components do not require alignment with physically meaningful properties and are heavily data-dependent \cite{pca_shortcomings}, thus making practical interpretation of PCA results difficult. To the best of our knowledge, the approaches proposed here have not been used so far.

Beside AMA and AMO, the rate of change of total AM about vertical axis (AMRCz) and total AM in spherical coordinates are also analyzed. Not only is AMRCz the derivative of AM, it also describes the net external torques acting on the system \cite{mechanics_textbook}. In this study, these net external torques are from the ground reaction forces (GRF) and moments (GRM). Since the main motion takes place about the vertical axis and involves direction change, it is of particular interest to analyze AMRCz. Total AM was also expressed in spherical coordinates to investigate its magnitude and direction in 3D. Force plate data are not covered in this paper since the data are too noisy for analysis.

\section*{Results}
\label{results}
The figures presented have time scaled for each phase via 1D interpolation. The average values presented are computed based on the true non-interpolated data. The Cartesian coordinates are defined based on the ready-pose assumed during Preparation (see Fig. \ref{fig:roundhousekick}, with X pointing to the right of the person, Y pointing forward, and Z pointing up. As mentioned, all AM variables presented are w.r.t. COM. Except for the total AM inclination angle and angular velocity, values are normalized to body mass and height squared. 

\subsection*{General Observations}
\label{results_genobs}
\begin{figure}[hbt!]
    \centering
    \includegraphics[height=0.9\textheight]{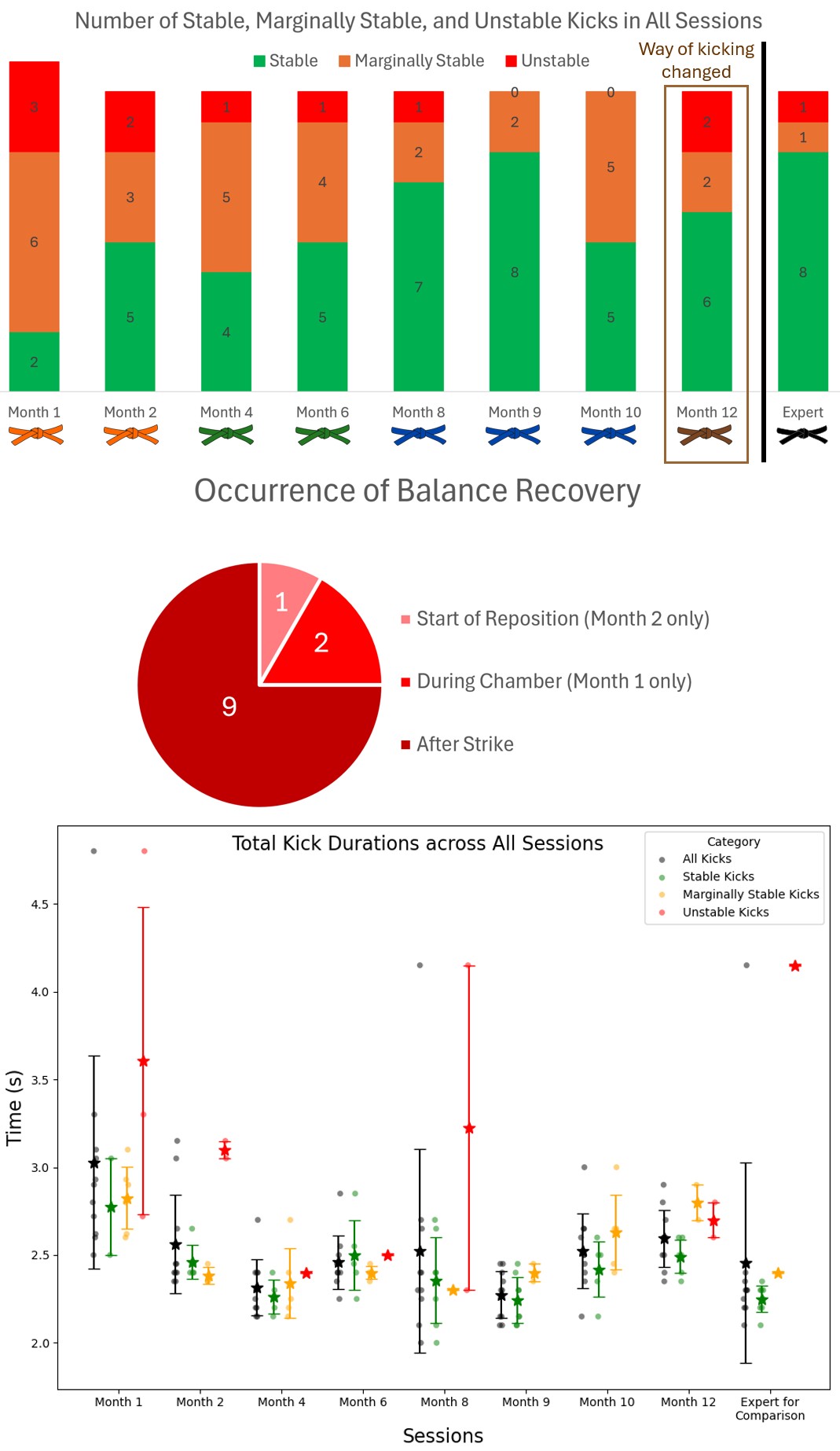}
    \caption{\textit{Top}: Summary of stable, marginally stable, and unstable kicks across all student sessions and expert's single session. \textit{Middle}: Breakdown of instability occurrence in unstable kicks. \textit{Bottom}: Mean (star) and standard deviation of all kick durations in all sessions.}
    \label{fig:kick_count_timeplot}
\end{figure}

Motion capture video data show that the student's roundhouse kicks in Months 1 and 2 are jerky. Starting from Month 4, not only do the student's kicks become smoother, the overall body posture is also less crouched. Based on a Karateka’s perspective on identifying stable kicks, there were a total of 50 stable, 29 marginally stable, and 12 unstable kicks of the 91 kicks performed. One kick in Month 1 (Kick 5) was incomplete due to an early failure point that prevented the kick from proceeding with the predefined five phases \textemdash leading up to Strike, the student lost balance from excessive counterclockwise rotation and had to take a step to regain balance. Fig. \ref{fig:kick_count_timeplot} shows the breakdown per session. Months 9 and 10 do not contain unstable kicks, and despite there being two unstable kicks in Month 12, an overall increase of the number of stable kicks is still observed throughout the year. Marginally stable kicks were not analyzed in this paper, but will be included in future analysis. For unstable kicks, the balance recovery strategy is observed immediately after Strike in nine kicks, during Chamber in two kicks, and at the start of Reposition in one kick. It can also be seen that the expert reacted to the instability differently \textemdash instead of salvaging the kick by forcefully rotating clockwise to retract, he stepped forward.

Throughout the year, the student’s average kick duration reduced from \(3.03s \pm 0.60\) to \(2.60s \pm 0.16\), showing an overall decrease of \(14\%\) and \(73\%\) in the mean and standard deviation respectively. Meanwhile, the expert’s average duration per kick is faster than the student's at\(2.46s \pm 0.57\). Unstable kicks have a longer average duration than stable and marginally stable kicks, except for Months 6 and 12. Details can be found in Fig. \ref{fig:kick_count_timeplot} and Supplementary Table S2 online.

\subsection*{Kick Leg Angular Momentum about Vertical Axis}
\label{results_AM}
\begin{figure*}[hbt!]
    \centering
    \includegraphics[height=0.8\textheight]{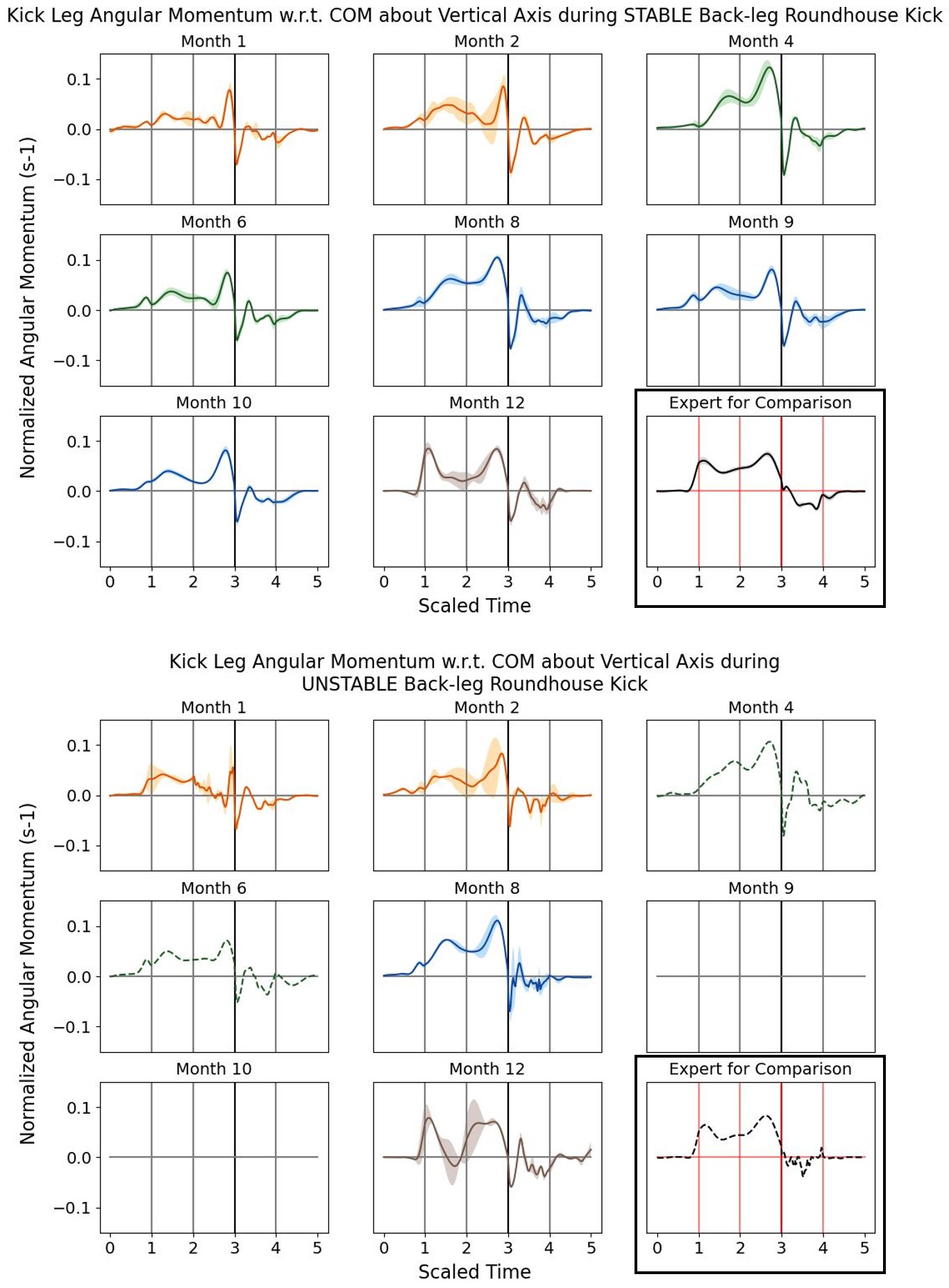}
    \caption{\textit{Top}: Kick leg AMz w.r.t. COM in stable kicks of all sessions. \textit{Bottom}: Kick leg AMz w.r.t. COM in unstable kicks of all sessions. Values are normalized to body mass and height squared. Mean and standard deviation are represented by solid line and shaded region respectively. Dotted trajectory means there is only one kick in the session. The darker vertical line represents the Strike instance.}
    \label{fig:rleg_AM_z_normalized}
\end{figure*}

According to the trajectory of the kick leg AM w.r.t. COM about the vertical axis (AMz) as illustrated in Fig. \ref{fig:rleg_AM_z_normalized}, an inflection point consistently occurs around Strike in all sessions. Only in Month 12 and the expert's session is there a larger initial peak occurring at the start of Chamber. Meanwhile, Month 12's unstable kicks show larger trajectory differences than its stable kicks, with AM being briefly negative and having a higher plateau during Extension. This is caused by an unstable kick reaching the negatives during Chamber and having an earlier peak during Extension, and another kick having a wider peak during Extension.

\subsection*{Rate of Change of Total Angular Momentum w.r.t. COM about Vertical Axis} 
\label{results_amrc}
\begin{figure*}[hbt!]
    \centering
    \includegraphics[height=0.8\textheight]{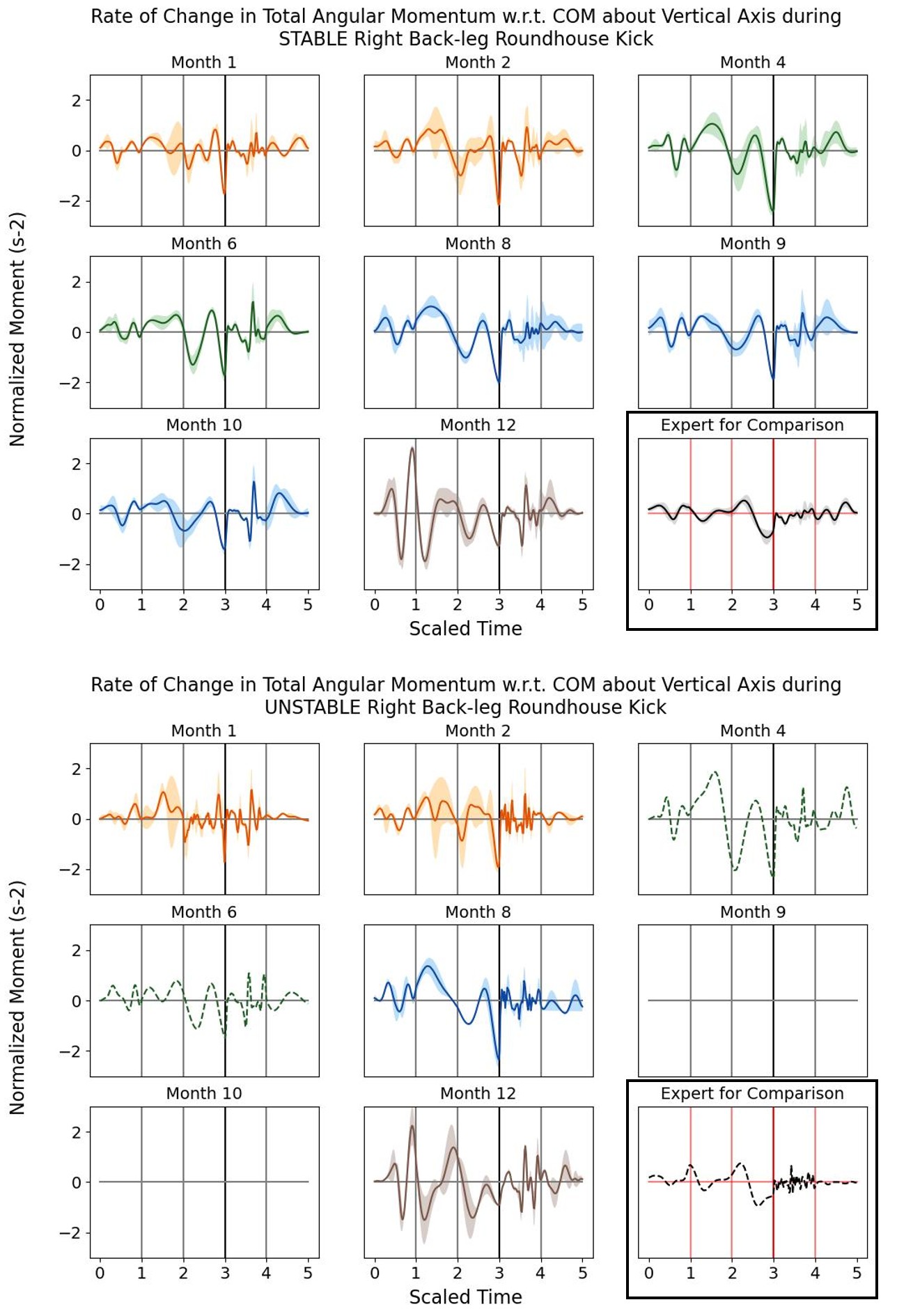}
    \caption{\textit{Top}: AMRCz w.r.t. COM in stable kicks of all sessions. \textit{Bottom}: AMRCz w.r.t. COM in unstable kicks of all sessions. Values are normalized to body mass and height squared. Positive means counterclockwise rotation, and negative means clockwise rotation. Mean and standard deviation are represented by solid line and shaded region respectively. Dotted trajectory means there is only one kick in the session. The darker vertical line represents the Strike instance.}
    \label{fig:amrcz_normalized}
\end{figure*}

Throughout each kick, AMRCz oscillates about zero and the negative peak occurring near Strike represents the amount of braking applied before Retraction. As shown in Fig. \ref{fig:amrcz_normalized} and excluding months without unstable kicks, 5/7 sessions' stable kicks have larger braking values than unstable kicks on average, with absolute intra-session differences ranging between \(0.10 s^{-2}\) (\(14.34 Nm\)) and \(0.31 s^{-2}\) (\(50.74 Nm\)) observed. Only a small difference of \(0.004 s^{-2} (0.66Nm)\) is observed in Month 1, and the unstable kicks in Month 8 instead have braking values larger by \(0.39 s^{-2} (62.12 Nm)\).

\subsection*{Total Angular Momentum Magnitude and Direction}
\label{AM_total}
\begin{figure*}[hbt!]
    \centering
    \includegraphics[height=0.8\textheight]{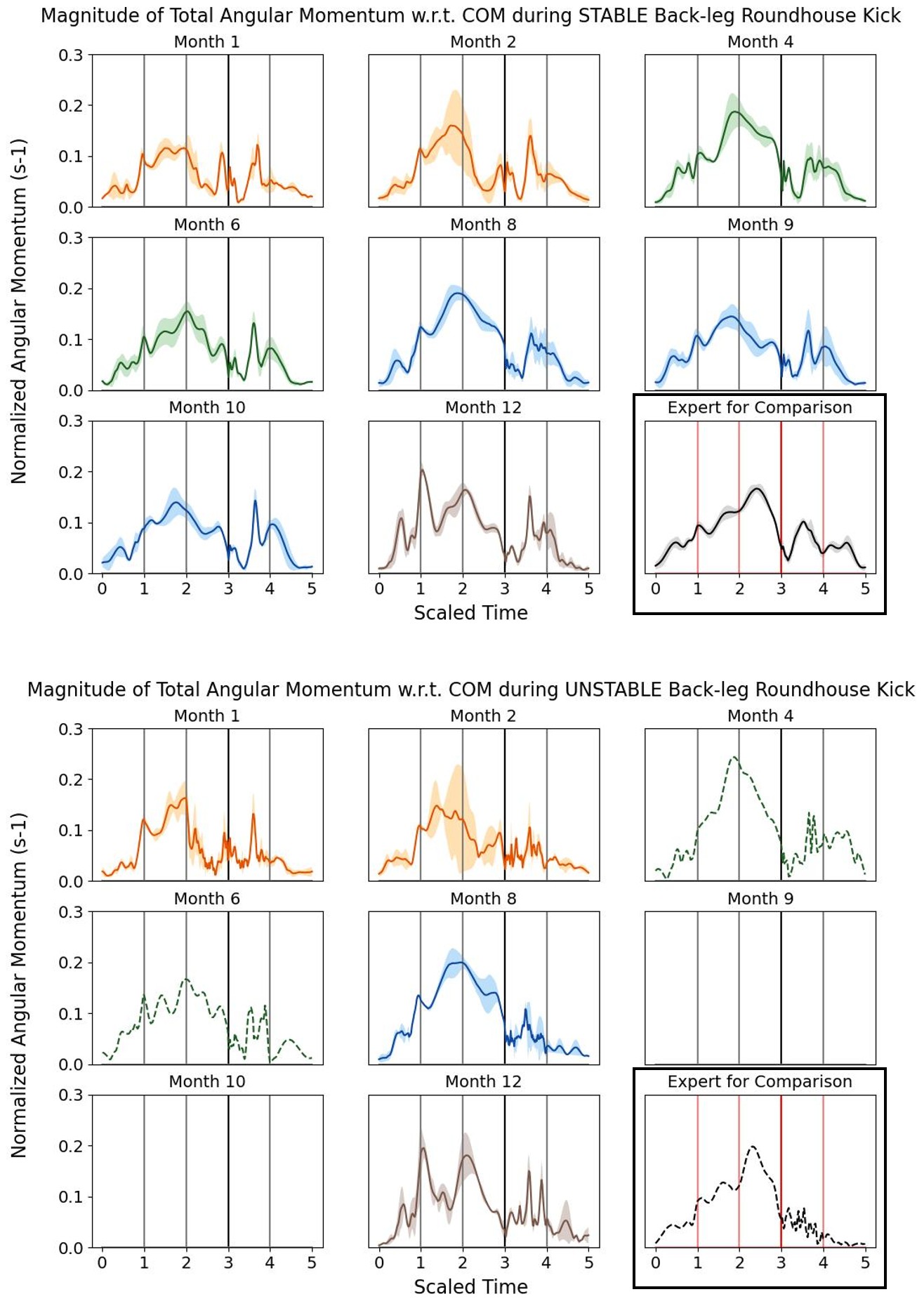}
    \caption{\textit{Top}: Total AM magnitude w.r.t. COM in stable kicks of all sessions. \textit{Bottom}: Total AM magnitude w.r.t. COM in unstable kicks of all sessions. Mean and standard deviation are represented by solid line and shaded region respectively. Dotted trajectory means there is only one kick in the session. The darker vertical line represents the Strike instance.}
    \label{fig:total_AM_normalized}
\end{figure*}

In all sessions with unstable kicks except Month 12, the unstable kicks consistently show a larger maximum magnitude than stable kicks of the corresponding sessions before Strike on average (see Fig. 
\ref{fig:total_AM_normalized}). The intra-session peak difference observed across sessions range from \(0.011 s^{-1} (1.7 kg m^2 s^{-1})\) to \(0.026 s^{-1} (7.6 kg m^2 s^{-1})\). In Month 12, no differences are observed. 

\begin{figure*}[hbt!]
    \centering
    \begin{subfigure}[b]{\textwidth}
    \centering
        \includegraphics[height=0.4\textheight]{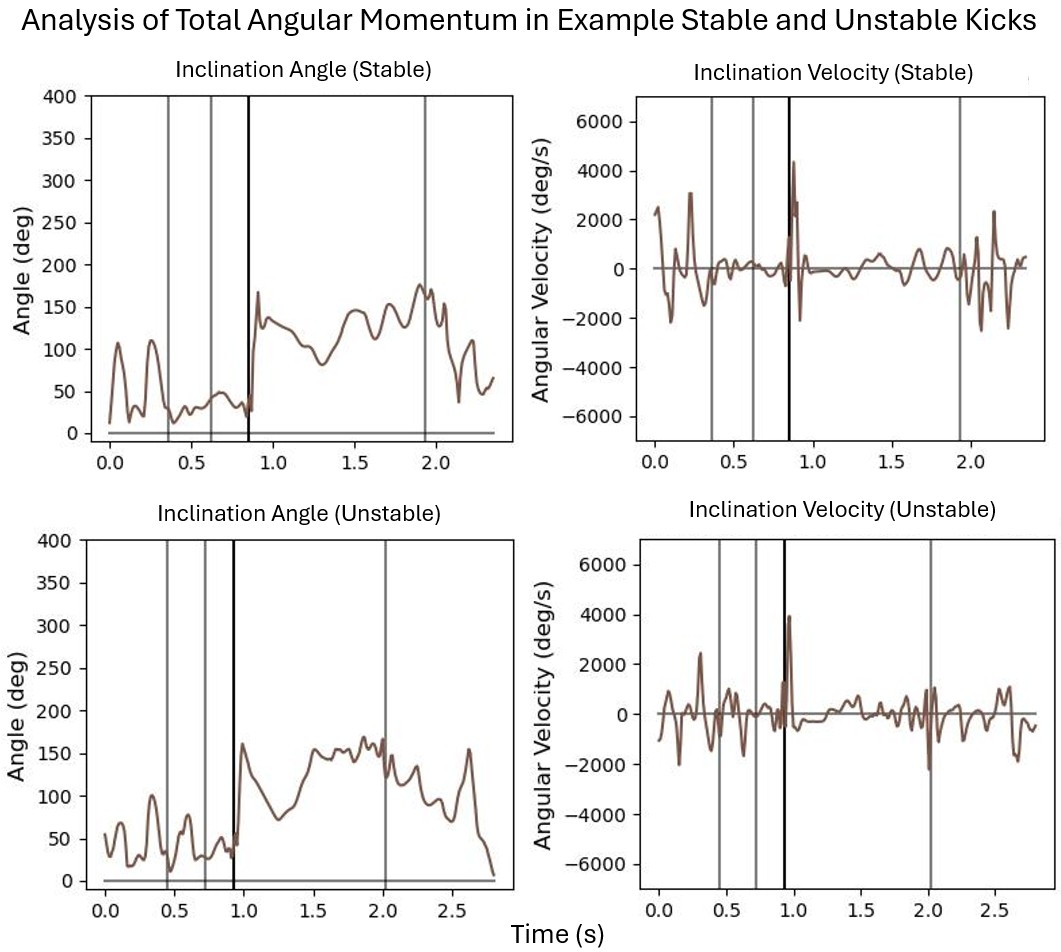}
        \caption{Inclination angle and velocity from example stable and unstable kicks. The darker vertical line represents the Strike instance.}
        \label{fig:inclination_plots}
    \end{subfigure}

    \vspace{0.3cm}
    \centering
    \begin{subfigure}[b]{\textwidth}
        \centering
        \includegraphics[height=0.45\textheight]{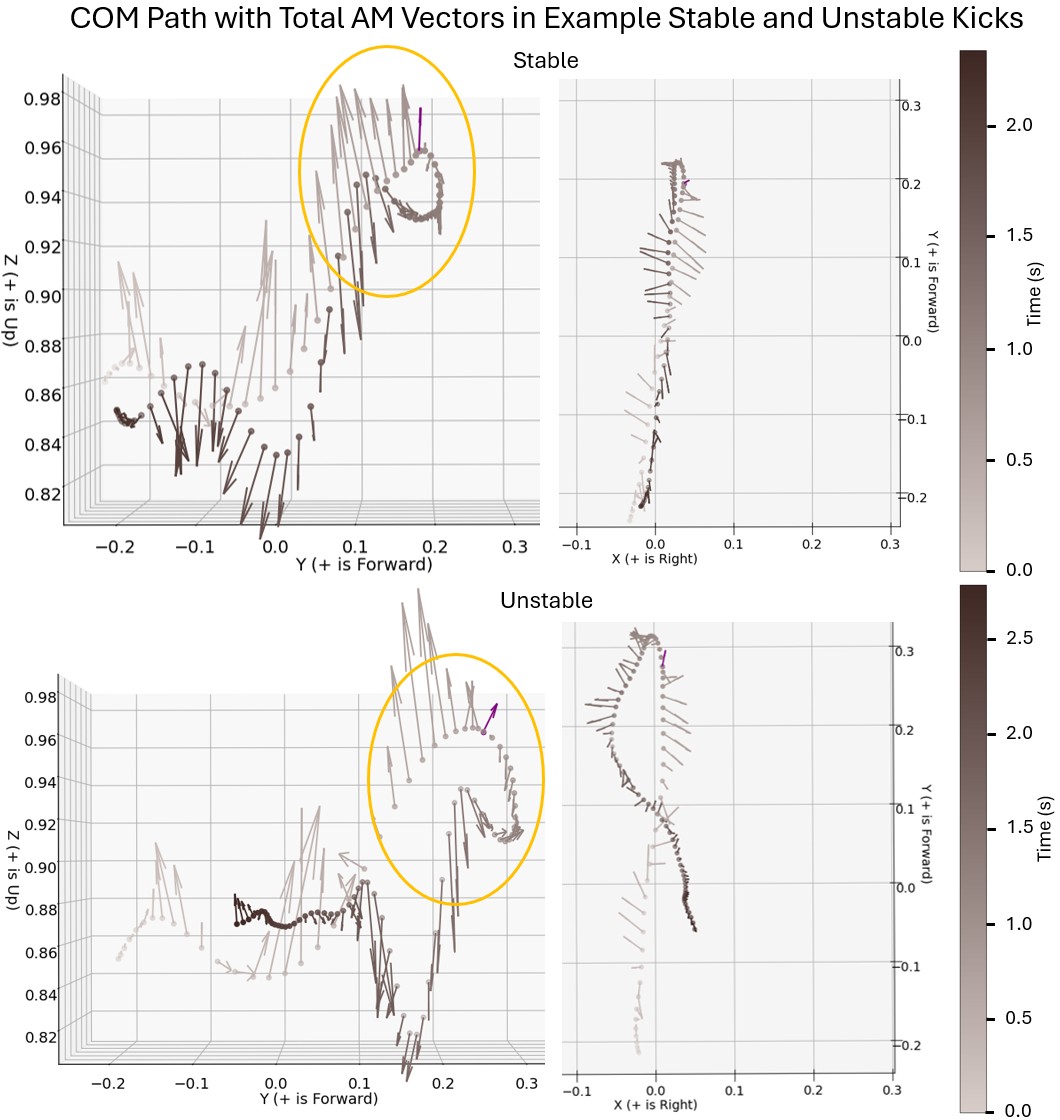}
        \caption{COM path with total AM vector in example stable and unstable kicks. A darker colour depicts further in time, and the purple dot and arrow denote the Strike instance. Yellow circle highlights the COM path behaviour around Strike.}
        \label{fig:comAMpathplots}
    \end{subfigure}
    \caption{Inclination angle, inclination velocity, and COM path with total AM vector in example stable and unstable kicks.}
    \label{fig:inclination_amcompath}
\end{figure*}

Fig. \ref{fig:inclination_amcompath} illustrates a typical example of the total AM vector's direction in a stable kick. Near Strike, a sharp change in both angles is observed. An inclination angle between \(0 ^\circ\) and \(90 ^\circ\) corresponds to a counterclockwise (positive) AMz rotation, while a value between \(90 ^\circ\) and \(180 ^\circ\) corresponds to a clockwise (negative) AMz rotation. Therefore, the sharp increase in the inclination angle around Strike stems from the vertical rotation's direction change. From the COM-AM path depiction, nonzero AM about the X axis (AMx) is observed. Positive AMx before Strike means that the person leans back, whereas negative AMx after Strike means that the person inclines forward when retracting the kick. The sharp changes in inclination angle is also reflected as a positive peak in the velocity plots. With the purple line representing the Strike instance and a darker colour progressing with time, it can be seen that the AM vector flips down abruptly after Strike, thereby also providing a visualization of the sudden increase in the inclination angle.

Fig. \ref{fig:inclination_amcompath} also shows an unstable example of a total AM vector's direction. Compared to the stable kick, the vector at Strike is tilted more forward and the peak inclination velocity near Strike is slightly smaller. The unstable kick's wider COM path during Retraction stems from the student's jump to the left in response to instability. Along the YZ-plane, the COM path in the unstable kick has a larger dip after Strike and a longer forward-backward range by approximately \(6.4 cm\).

\subsection*{Angular Momentum Aligned Component of Kick Leg} 
\label{results_ama}
\begin{figure*}
    \centering
    \includegraphics[height=0.9\textheight]{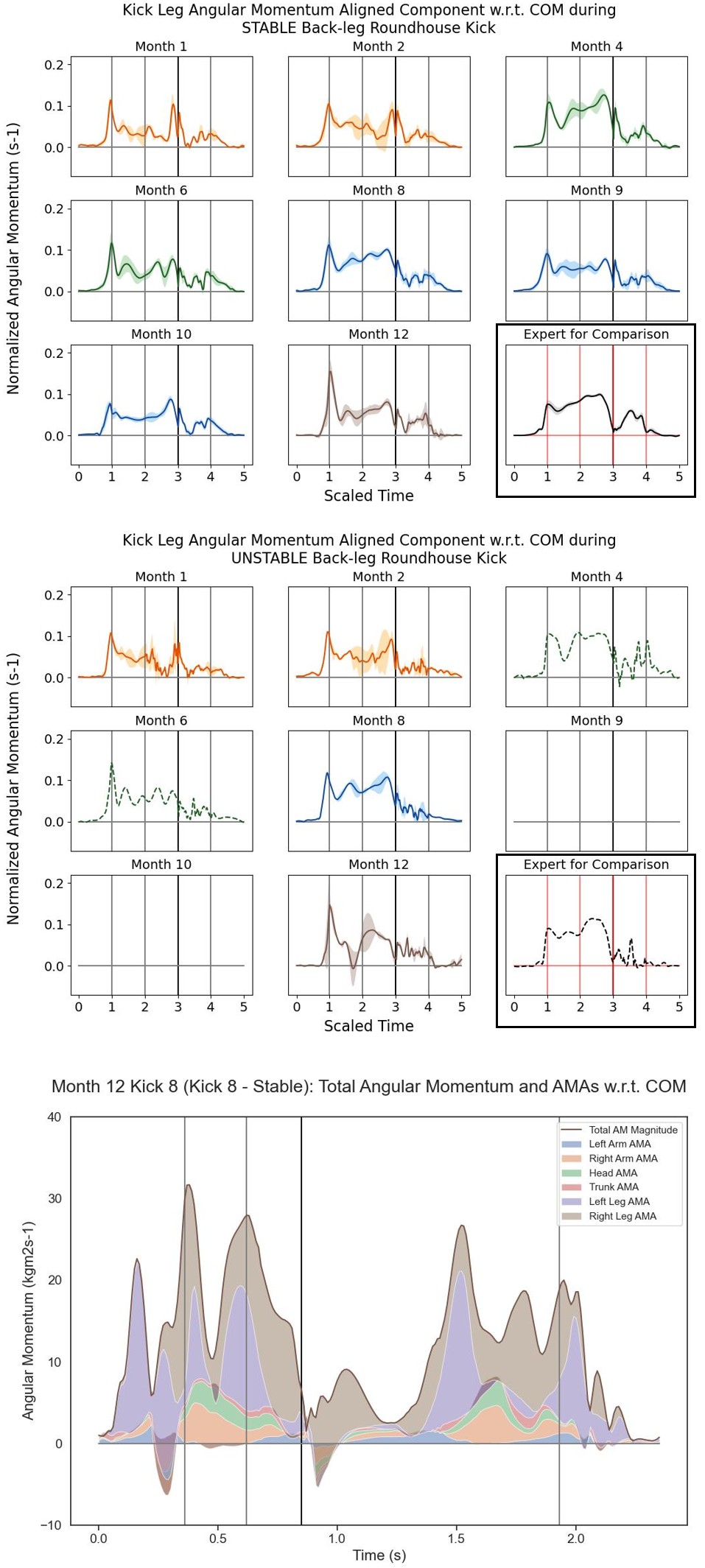}
    \caption{\textit{Top}: Kick leg AMA in stable kicks of all sessions. \textit{Middle}: Kick leg AMA in unstable kicks of all sessions. Mean and standard deviation are represented by solid line and shaded region respectively. Dotted trajectory means there is only one kick in the session. \textit{Bottom}: An example from a stable kick illustrating the total AM magnitude (solid line) and breakdown of body part AMAs (shaded regions). The darker vertical line represents the Strike instance.}
    \label{fig:amcom_ama_eg_s8k8_normalized}
\end{figure*}

Fig. \ref{fig:amcom_ama_eg_s8k8_normalized} depicts the body part AMA results. The kick leg AMA of the student’s stable kicks always shows a distinct peak at the end of Preparation, whereas in the expert's stable kicks, this peak is less distinct than the student's and there is a steady increase until it reaches the global peak during Extension. Regarding unstable kicks, the expert's has a global peak \(13.3\%\) larger than his stable kicks on average, and one of the student's unstable kicks in Month 12 has a negative peak of \(-0.05 s^{-1} (-8.31 kgm^2s^{-1})\) during Chamber.

\subsection*{Angular Momentum Orthogonal Component of Kick Leg} 
\label{results_amo}
\begin{figure*}[hbt!]
    \centering
    \includegraphics[height=0.8\textheight]{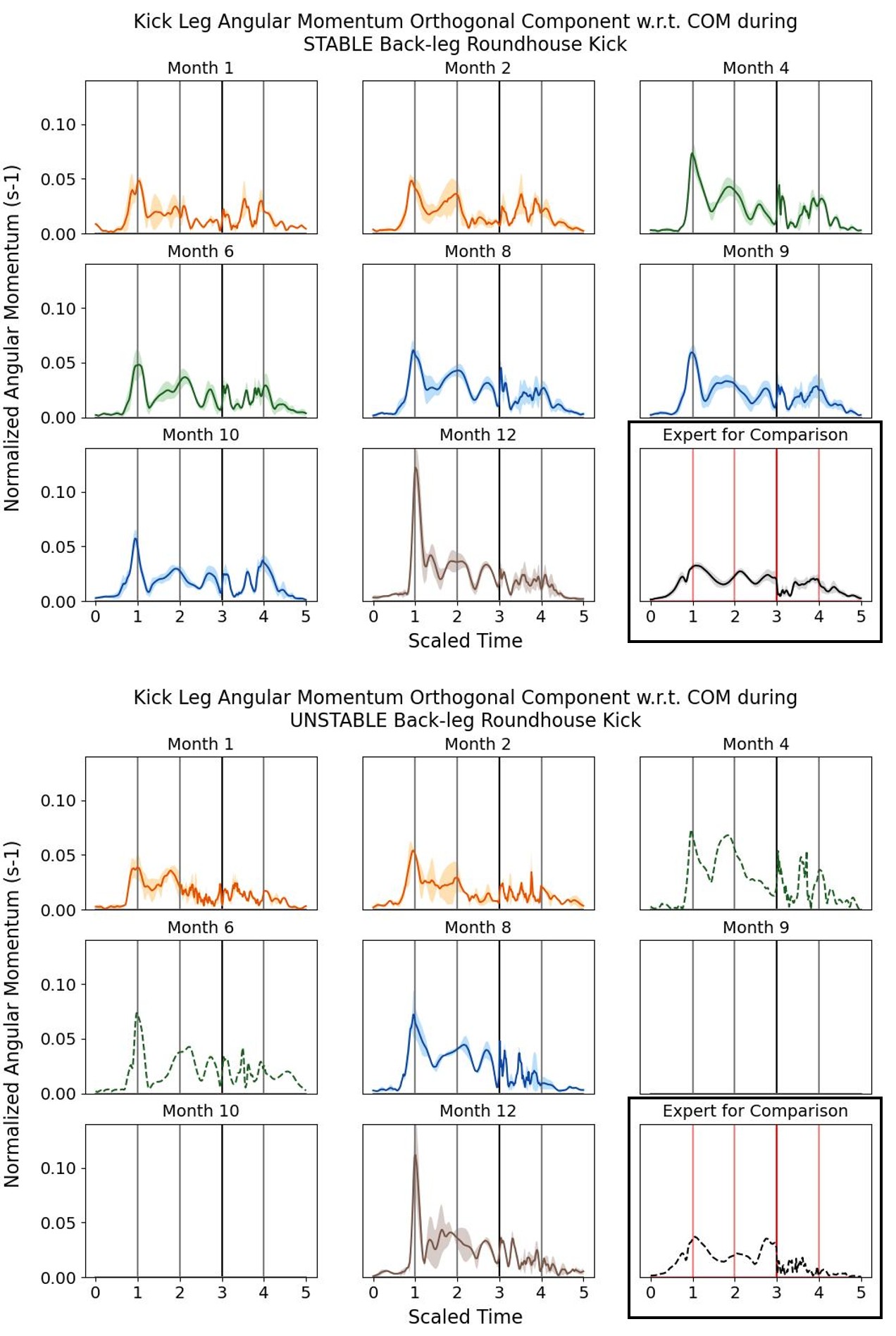}
    \caption{\textit{Top}: Kick leg AMO in stable kicks of all sessions. \textit{Bottom}: Kick leg AMO in unstable. kicks of all sessions. Values are normalized to body mass and height squared. Mean and standard deviation are represented by solid line and shaded region respectively. Dotted trajectory means there is only one kick in the session. The darker vertical line represents the Strike instance.}
    \label{fig:rleg_amo_normalized}
\end{figure*}

Similar to kick leg AMA, the student has a distinct kick leg AMO peak occurring at the end of Preparation (see Fig. \ref{fig:rleg_amo_normalized}). Throughout the year, the student's peak kick leg AMO increases, with Month 12's stable kicks having the highest normalized average peak value of \(0.13 s^{-1}\) (\(20.51 Nm\)). Meanwhile, the expert’s stable kicks have a much smaller peak averaging at \(0.032 s^{-1} (9.50 Nm)\) at the end of Preparation. As for unstable kicks, Month 4's peak during Chamber is \(40.9\%\) larger than its stable kicks, whereas the expert's peak shortly before Strike is \(35.2\%\) larger than its stable kicks.

\clearpage
\section*{Discussion}
\label{discussion}
The purpose of this longitudinal study is to understand AM management of Karate roundhouse kicks using different AM variables and the newly-proposed measures while analyzing kick stability in the form of improvement tracking over one year. In this section, the hypotheses are first addressed, followed by additional discussion related to the results and limitations.

\textbf{Hypothesis 1.} The motion incoherence observed in Months 1 and 2 is expected, since Karate beginners are taught to prioritize proper technique over speed, which leads to the occurrence of kinematic and visual distinction between phases. Smoother kick execution and larger kick leg AMA, AMO, and AMRCz ranges in later months also indicate higher comfort levels of performing the kick dynamically. While the larger AMA, AMO, and AMRCz amplitudes in the student's later months reveal her capability in performing the kick in a more dynamic fashion, the expert's smaller amplitudes show the possibility of performing the back-leg roundhouse kick dynamically with less drastic footwork and reduced AM components from the kick leg, all while preserving Karate technique and being stable. Therefore, the first hypothesis is only partially supported.

\textbf{Hypothesis 2}. Recall that among all sessions with unstable kicks, 6/7 sessions follow the trend of unstable kicks exerting higher peak total AM before Strike. This observation makes sense because the person is responsible for stopping on their own without a kicking pad, and if there is too much AM before Strike, it becomes challenging to retract stably. Therefore, the second hypothesis is supported.

\textbf{Hypothesis 3}. In the beginning of all kicks, a peak is observed in the kick leg AMA and AMO, illustrating the fact that the kick leg has a large influence on positively contributing to the total AM as well as changing its direction. Kick leg AMA and AMO can also potentially suggest instances that cause or are an outcome of instability in three unstable kicks. For the expert, the instability is caused by too much kick leg AMA before Strike and the subsequent kick leg AMO peak could be attempt to reduce his total AM for stability reasons. In Month 4, the larger kick leg AMO before Strike suggests that the instability can potentially be caused by the kick leg trying to change too much of the total AM's direction. However, the kick leg AMO threshold between stable and unstable kicks can differ across sessions and also depend on the way of kicking, hence the third hypothesis is not fully supported.

\textbf{Hypothesis 4}. In Month 12, one of the unstable kicks' negative kick leg AMA before Strike could be the cause of instability. Since the kick leg and fullbody are rotating in the same direction, and given the fast and dynamic nature of the motion, the kick leg AM going in the opposite direction of total AM could hinder postural control. Therefore, the fourth hypothesis is supported.

Faster kicks alone generally do not mean better kicks, though the expert's shorter kick duration demonstrates the possibility of performing the Karate back-leg roundhouse kick faster without compromising stability and technique. Unstable kicks have longer duration on average due to recovery time from instability. The student's reduced standard deviation values with an overall increase in the number of stable kicks throughout the year indicate improved consistency and kick control.

For the kick leg AMz, the small value at Strike occurs when the person stops and turns around to retract the kick with their own efforts. The student's kick leg AMz trajectory in Month 12 being similar to the expert's demonstrate a shift in paradigm of learning, though the kick leg AMA and AMO trajectories reveal that the student is still kicking differently than the expert since the student's kick leg AM has a much larger influence on changing the total AM versus the expert's. Nevertheless, the sudden trajectory shape change in Month 12 could still indicate that roundhouse kick mastery is a staged process rather than a gradual one.

Regarding the total AM direction, the COM-AM path visualization in the stable example shows that AM about the horizontal axes are also utilized during retraction. It is unsure whether the subtle coordination between the backward/forward leaning and forward/backward COM translation improves stability, so more research must be conducted. As for the unstable example, the differences in COM path and AM vector directions around and after Strike are likely a response to instability, rather than the cause of instability. Meanwhile, when AMz crosses zero during Strike, it creates an artifact that causes the inclination angular velocity to have large peaks. The original intent of analyzing this variable is to investigate a different perspective on how fast the total AM vector points downwards when the person transitions from counterclockwise rotation to clockwise rotation. Although the peak values are numerically correct, it is naturally sensitive to noise from zero-crossings and thus yields low interpretability when it comes to systems changing directions about the vertical axis. The angle change can be large, but one must remember that these plots do not convey any information about the magnitude of the AM, which is further proven with the COM-AM path visualization. Even though analyzing total AM direction along with magnitude can provide more insight on performance and stability, the current method of portraying total AM in spherical coordinates analysis yields angular information that is sensitive to noise. Therefore, it is recommended to apply or develop additional processing methods for these variables to improve data interpretability.

As mentioned, AMRCz describes the supporting foot's rotational acceleration and braking during each kick, since it describes how the GRF and GRM act on the fullbody COM. Despite the general trend observed where stable kicks have larger negative AMRCz peaks prior to Retraction, Month 8 does not follow this trend. This makes sense because too much braking can also cause instability due to difficulty in retracting kick when a person comes close to a full halt. Moreover, the trajectories from all sessions also reveal an interesting phenomenon in the braking strategy, where despite only one direction change in the kick, the multiple oscillations about zero reveal that the supporting foot constantly adjusts how the GRF and GRM act on the person.

Even though the AM measures and variables presented demonstrate the potential of describing body part AM influence on the total AM, the current analysis performed have limitations. Since kicks can be unstable for different reasons and particularly given the motion's dynamic nature, for the other unstable kicks that have not been mentioned, it is difficult to identify how these variables behave differently in the unstable kicks, as well as when the values are caused by the nature of the motion and when the values are related to preventing instability. Therefore, if one were to analyze stability of dynamic motions, it is recommended to not only analyze the variables separately as done in this paper, but also to combine all these measures with kinematic, time, and potentially center-of-pressure data to develop a novel criterion. An important note is that one must be wary of turning AMA and AMO into a contribution by dividing them by total AM, since this runs into the risk of diving by zero when total AM is close to zero. Although one can generate additional conditions to avoid numerical artifacts of dividing by zero, this can complicate the data interpretation.

\section*{Conclusion}
\label{conclusion}
Unstable roundhouse kicks can occur for different reasons, as is the case for instability in ADL. Therefore, it can be challenging to determine dynamic stability purely based on AM-related threshold values. As a first of many steps towards developing a novel AM-based stability criterion for anthropomorphic systems performing dynamic motions, this paper is an in-depth investigation of AM behaviour in the Karate back-leg roundhouse kick, which is fast and dynamic in all directions. Results show that the kick leg AMz can inform the student's shift in paradigm of learning, yet the kick leg AMA and AMO can further describe how differently the student is kicking compared to the expert. Kick leg AMA and AMO show some indication of instability in a few unstable kicks, though more research must be conducted. In general, unstable kicks have higher total AM magnitude before Strike and less braking moment after Strike. Despite limitations, AMA and AMO show potential to be used as performance measures. Not only do they convey 3D information with simple calculations, they can also describe the underlying AM behaviour of each body part in relation to total AM. However, if one were to create stability criteria with them, it is recommended to investigate their coordination with time, kinematic, and kinetic variables.

\section*{Data availability statement}
Joint angle data for all trials are available as Supplementary Data. AM plots of each body part are also available in the Supplementary Note.

\bibliography{references}

\section*{Acknowledgements}
This research is possible thanks to funding from the Hector II Foundation and Canada Excellence Research Chair. Special thanks to Marko Ackermann and Jonas Grosse Sundrup for interesting discussions, and Maximillian Schik for improving the vicon-biorbd pipeline.

\section*{Author contributions statement}
J.C.L.L., C.M., and J.F.-S.L. conceived and conducted the experiment. J.C.L.L. and K.M. developed the proposed AM measures and analyzed the results. C.M. provided Karate expertise. J.C.L.L. curated the data and wrote the original draft. K.M. acquired funding, provided supervision, reviewed and curated the manuscript. All authors reviewed the manuscript.

\section*{Additional information}
\textbf{Competing interests} The authors declare no competing interests.

\noindent \textbf{Supplementary Information} The online version contains supplementary material available at (link from Scientific Reports).

\end{document}